\pdfoutput=1

\documentclass[11pt]{article}

\usepackage{acl}
\usepackage{times}
\usepackage{latexsym}

\usepackage[T1]{fontenc}

\usepackage[utf8]{inputenc}

\usepackage{microtype}

\usepackage{times}
\usepackage{adjustbox}
\usepackage{hyperref}
\usepackage{comment}
\usepackage{latexsym}
\usepackage{booktabs}
\usepackage{graphicx}
\graphicspath{ {./images/} }
\usepackage{subcaption}
\usepackage{amsmath}
\usepackage{siunitx}
\usepackage{array,multirow,graphicx}
\usepackage[T1]{fontenc}
\usepackage{enumitem}
\usepackage[capitalise]{cleveref}
\usepackage{pmboxdraw}

\usepackage{multirow}
\usepackage{cleveref}
\usepackage{textcomp}
\usepackage[ruled,vlined]{algorithm2e}
\crefformat{section}{\S#2#1#3}

\usepackage[T1]{fontenc}

\usepackage[utf8]{inputenc}

\newcolumntype{C}{>{\centering\arraybackslash}m{0.3\linewidth}}
\newcolumntype{D}{>{\centering\arraybackslash}m{0.13\linewidth}}
\newcolumntype{E}{>{\centering\arraybackslash}m{0.16\linewidth}}
\newcolumntype{f}{>{\centering\arraybackslash}m{0.41\linewidth}}
\newcolumntype{G}{>{\centering\arraybackslash}m{0.20\linewidth}}
\newcolumntype{H}{>{\centering\arraybackslash}m{0.15\linewidth}}
\newcolumntype{I}{>{\centering\arraybackslash}m{0.14\linewidth}}
\newcolumntype{J}{>{\centering\arraybackslash}m{0.17\linewidth}}
\newcolumntype{K}{>{\centering\arraybackslash}m{0.18\linewidth}}
\newcolumntype{L}{>{\centering\arraybackslash}m{0.11\linewidth}}
\newcolumntype{M}{>{\centering\arraybackslash}m{0.1\linewidth}}
\newcolumntype{N}{>{\centering\arraybackslash}m{0.19\linewidth}}
\newcolumntype{W}{>{\centering\arraybackslash}m{0.07\linewidth}}
\newcolumntype{Z}{>{\centering\arraybackslash}m{0.21\linewidth}}
\usepackage{microtype}

\usepackage{threeparttable}

\title{Low-Resource Machine Translation Training Curriculum Fit for Low-Resource Languages}

\author{Garry Kuwanto\thanks{\ \ Contributed equally} \\ Institut Teknologi Bandung \\  \texttt{gkuwanto@gmail.com} \And Afra Feyza Akyürek\footnotemark[1] \\ Boston University \\ \texttt{akyurek@bu.edu} \And Isidora Chara Tourni\footnotemark[1]\\ Boston University \\ \texttt{isidora@bu.edu}
         \AND
         Siyang Li\footnotemark[1] \\ Boston University \\ \texttt{siyangli@bu.edu} \And Alexander Gregory Jones\\Dartmouth College \\ \And Derry Wijaya \\ Boston University\\ \texttt{wijaya@bu.edu}}

\begin{document}
\maketitle
\begin{abstract}
We conduct an empirical study of neural machine translation (NMT) for truly low-resource languages,
and propose \textit{a training curriculum} fit for cases 
when \textit{both} parallel training data and compute resource are lacking, reflecting the reality of most of the world's languages and the researchers working on these languages. 
Previously, unsupervised NMT, which employs back-translation (BT) and auto-encoding (AE) tasks has been shown barren for low-resource languages. We demonstrate that leveraging comparable data and code-switching as weak supervision, combined with BT and AE objectives, result in remarkable improvements for low-resource languages even when using only modest compute resources.  
The training curriculum proposed in this work achieves BLEU scores that improve over \textit{supervised} NMT trained on the same backbone architecture by +12.2 BLEU for English$\rightarrow$Gujarati and +3.7 BLEU for English$\rightarrow$Kazakh, showcasing the potential of \textit{weakly-supervised} NMT for the low-resource languages. When trained on  \textit{supervised} data, our training curriculum achieves a new state-of-the-art result on the Somali dataset (BLEU of 29.3 for Somali$\rightarrow$English). We also observe that adding more time and GPUs to training can further improve performance, which underscores the importance of reporting compute resource usage in MT research. 
\end{abstract}

\section{Introduction}

Approaches in low-resource NMT research have seen significant progress recently. Facilitating translation for extremely low-resource languages with no parallel training data is a growing number of unsupervised NMT approaches that use techniques such as back-translation and auto-encoding \citep{lample2018unsupervised,artetxe2017unsupervised,sen2019multilinguall,artetxe2019effective,sun2019unsupervised} pre-trained on a growing number of massively multilingual language models (LMs). 
With large LMs prevalent in NLP, such as BERT and GPT-3, we can presumably achieve good low-resource NMT performance if we can pre-train the LMs on a large amount of data: hundreds of millions of sentences (including monolingual and parallel data from related higher-resource languages), on large compute resources: hundreds of large memory GPUs, for a long period of time (weeks) \cite{liu2020multilingual}. 

However, the \textit{assumption} of abundant monolingual data, available parallel data to/from the target language or in related higher-resource languages such as Russian for Kazakh \cite{zoph2016transfer,nguyen2017transfer,dabre2017empirical,kocmi2018trivial}, or the assumption of abundant compute resources, remains unsubstantiated when it comes to NMT for truly low-resource languages--
making the presumed approach inapplicable to hundreds of the so-called \emph{left-behind} languages \cite{joshi-etal-2020-state}. Moreover, multilingual LMs, due to their limited capacity and large differences in pre-training data sizes, have large performance gaps between high- and low-resource languages \cite{wu-dredze-2020-languages} and have difficulties transferring to languages that do not exist in their pre-training \cite{muller2020being, pfeiffer2020mad}. The compute and environmental cost of re-training these LMs to include a language of interest is prohibitively large \cite{strubell-etal-2019-energy}. 

Given the challenges of working with truly low-resource languages, most works on low-resource NMT have focused on \textit{simulating} low-resource scenarios either for high-resource languages (e.g., German) for which low-resource methods are redundant, or for medium-resource and relatively similar to English ones (e.g., Romanian). 
When they are finally applied to truly low-resource languages that are typologically distant from English (e.g., Gujarati), 
the best performance of unsupervised NMT 
is spectacularly low--BLEU scores of less than 1 \cite{kim2020and,liu2020multilingual}, giving an impression of its ``uselessness''. In this work, we show that there are comparable data such as Wikipedia that can be easily mined to provide weak supervision to low-resource NMT, and that  combining training on these comparable data with pre-training on the unsupervised NMT task 
is \textit{not useless}, on the contrary, an \textit{integral} part of the training curriculum effective for truly low-resource languages. Along with code-switching pre-training, they together outperform their individual scores. The proposed  curriculum can also be conducted with supervised data and boosts \textit{supervised} NMT performance from English to these languages. These results are promising given 
the overwhelming need for automatic translation \emph{from} English \emph{to} other languages, as information in the former is ample compared to the latter \cite{nekoto2020participatory}. 

In this paper, among the low-resource languages in MT for which there are test corpora to evaluate MT models on, we focus our attention to three: Gujarati (gu), Somali (so), the \textit{scraping-by} languages---according to \citet{joshi-etal-2020-state}---for which \textit{some} monolingual data are available, and Kazakh (kk), the \textit{rising-star} language, for which \textit{more} monolingual data is available. These are distant languages that are diverse in terms of script, morphological complexity and word order from English (en) and for which unsupervised NMT has either never been explored (for so) or has been shown to perform poorly i.e., 0.6 and 0.8 BLEU for en$\rightarrow$gu and en$\rightarrow$kk \cite{kim2020and}. 

Wikipedia, our source of comparable data in this work, has over 300 languages, both high and low-resource, allowing our approaches here to scale to other languages. We mine comparable sentences from Gujarati, Somali, and Kazakh Wikipedia, that are linked to their corresponding English articles 
by using a dictionary---Panlex \cite{kamholz2014panlex}---which are available in 5,700 languages) to mine sentences with lexical overlap. 
We also use the dictionary to word-translate monolingual data (news data) and obtain code-switched sentences, 
which we use to boost pre-training of our bilingual LM. 
Although we have strived to use resources that are publicly available for many languages (Panlex, Wikipedia and news), we understand that these may still not apply to many of the world’s languages. 


Further, unless otherwise indicated, we assume limited access to compute resource and restrict our compute expenses by the average monthly income for the language speaking region\footnote{For each language, we use the average monthly income estimated in  \href{https://www.numbeo.com/cost-of-living}{https://www.numbeo.com/cost-of-living}, and calculate how many hours of 1 GPU (32GB) training time we can get for 1 month of income using the AWS EC2 rate \href{https://calculator.aws}{https://calculator.aws}. This comes down to 40, 60, and 72 hours of training for gu, so, and kk respectively}. Our assumption of limited access to compute resource is realistic and driven by personal experiences and previous works that have observed how indeterminate access to a large number of GPUs is unrealistic beyond large industrial/academic
environments \cite{ahmed2020democratization}. By using limited compute resources, we also 
hope to shed light on the true state of low-resource NMT (and the researchers working on it) when both data \textit{and} access to compute resource are lacking. We observe in our experiments that without this  constraint, 
adding even a small compute power from 1 GPU (32GB) to 4 GPUs (32GB each) 
\textit{or} training for a longer period of time can already improve NMT performance. 
Since previous works have shown that increasing compute resources comes with a high environmental cost \cite{strubell-etal-2019-energy, timnit2021stochasticparrot}, we echo their suggestions 
to factor in the compute resources when reporting our results. 


To summarize, our contribution 
within the context of low-resource languages is to introduce of an effective training curriculum for NMT with no dependency on available parallel data in the language or from related languages, by leveraging comparable data and monolingual data through code switching and unsupervised training.  
We extensively compare our training curriculum to other strategies, unsupervised and supervised. 
Finally, without supervised training data, 
we achieve remarkable improvements  from English to these languages, with improvements up to +14.2 BLEU points from unsupervised results and up to +12.2 BLEU points from supervised results that are built on the same backbone architecture as ours. 


\section{Related Work}

\citet{joshi-etal-2020-state} argue that despite the rapid progress in language technologies, research efforts in NLP have only incorporated about 6\% of all 7000 world languages. In their study, the authors develop a taxonomy of six categories depending on the data available in each language (labeled and unlabeled), and study extensively their resource disparities, and representation in NLP conferences. 
Their analysis highlights that NLP methods and venues need to further focus on under-explored and typologically diverse languages. This motivates our choice of languages in this paper since so, gu, and kk are low-resource and typologically diverse languages that are also under-explored in NMT. 

Aside from the gap in available data, a 
growing phenomenon is the “compute divide”  
caused by the large computational requirements in GPU usage 
and the researchers’ unequal access to computing resources \cite{ahmed2020democratization,strubell-etal-2019-energy}. We believe that a discussion of compute constraints should always be included in the study of low-resource NMT; 
since for many of these languages the lack of compute resource, infrastructure, and time constraints can hinder communities in low-resourced societies from working and publishing on these languages; and can render the use of techniques developed in high-resourced societies inapplicable in this low-resource setting \cite{joshi-etal-2020-state,nekoto2020participatory}. 

When parallel data is scarce, unsupervised NMT can play a crucial role. 
However, previous works have only focused on high-resource and/or similar-to-English languages. Recent works have also questioned the universal usefulness of unsupervised NMT and showed its poor results for low-resource languages \cite{kim2020and,marchisio2020does}. They reason that this is because  factors that are important for good unsupervised NMT such as linguistic similarity
, domain proximity along with size and quality of the monolingual corpora are hard to satisfy in the case of low-resource languages. 
In this work we show that a training pipeline involving an unsupervised MT training \cite{lample2018unsupervised} followed by training on comparable text improves performance significantly. 


There is a large body of work in mining comparable i.e., \textit{pseudo}-parallel sentences \citep{munteanu2004improved,munteanu2006extracting,zweigenbaum2017overview,guo2018effective,grover2017bilingual,schwenk2018filtering,hangya2018unsupervised,hangya2019unsupervised,wu2019machine, resnik2003web}; yet most approaches have not been widely applied in the low-resource scenarios. 
Some of the recent works like CCMatrix or WikiMatrix also rely on supervised systems trained on parallel data \cite{schwenk2019wikimatrix,schwenk2019ccmatrix,pourdamghani2019translating} and/or require extensive compute resource 
\cite{tran2020cross}. 

In this paper, we use a mining approach that is based on lexical overlap similar to STACC \cite{azpeitia2018extracting}. Our lightweight mining however, is not challenging other approaches, indeed can be substituted by them, but aims to be used as a starting point by a practitioner who wants to train an MT system for a low-resource language with as little resource as possible. 
Our main contribution is not on comparable corpus mining, rather on how and when we introduce the comparable data into the training of the model. For completeness of our work, we compare our mining method to state-of-the-art (SOTA) unsupervised mining (\cref{subsec:mining-eval}), because our training curriculum does not assume any parallel data. Using other approaches based on supervised mining methods trained with parallel data i.e., CCMatrix or WikiMatrix, would be contrary to our non-reliance on parallel data.

This most recent unsupervised bitext extraction method that we compare to utilizes multilingual contextual embeddings retrieved from mBERT \citet{keung2020unsupervised}. 
Unlike this approach that requires multilingual model and expensive computation of contextual embedding similarity between sentences, we employ a simpler method of using bilingual dictionaries (Panlex) to mine sentences with lexical overlap, similar to STACC \cite{azpeitia2018extracting} and \citet{resnik2003web, ma1999bits}. 
In comparison to STACC, which uses top-k word translations from parallel sentences, we only use top-1 translations from our dictionary, 
which also relies only on a single translation direction (XX$\rightarrow$EN).

Regarding code-switching in MT, our paper uses code-switching during the LM training to improve the cross-lingual alignment and downstream MT, hence differs from \citet{yang-etal-2020-csp}, who conduct code-switching during the MT model training. Our work is more similar to \citet{yang2020alternating}, however we use only monolingual sentences and evaluate on low-resource languages, while they use parallel sentences to create code-switched data and evaluate on high resource languages. 


\section{Method}
\label{sec:model}

\subsection{MT Training Curriculum}

Since transformer-based architectures have proven successful in NMT numerous times \cite{barrault2019findings}, for all of our experiments we use XLM \cite{conneau2019cross} as our backbone architecture. Our proposed training curriculum starts with pre-training a bilingual Language Model (LM) using the Masked Language Model (MLM) objective \cite{devlin-etal-2019-bert} on the monolingual corpora of two languages (e.g. so and en for en-so MT). To encourage the LM to align English and the foreign language embeddings, we also propose adding a third ``language'', the corpus of which consists of code-switched sentences containing words in English and the foreign language (\cref{data:cs}). With the MLM objective, to predict a masked English word, the model can attend to both the English and foreign language words in the code-switched sentence, and vice versa. This is akin to the Translation Language Modeling (TLM) objective in XLM, 
but without the need of parallel sentences. 


After pre-training the LM, we further pre-train the model on an unsupervised NMT task,  
following the setup recommended in \citet{conneau2019cross}, where both encoder and decoder are initialized using the same pre-trained LM encoder block. For unsupervised NMT, we use back-translation (\emph{BT}) and denoising auto-encoding (\emph{AE}) losses \cite{lample2018unsupervised}, using the same monolingual data as in LM pre-training, delineated in \cref{sec:monodata}. We follow this unsupervised \emph{BT+AE} pre-training with \emph{BT+MT$_c$}, where \emph{MT$_c$} stands for supervised machine translation objective for which we use mined comparable data.

\begin{table*}[ht]
\centering
\small
\begin{tabular}{c H H H H H}
\toprule
\textbf{} & \textbf{Wikipedia} & \textbf{WMT (2018/2019)} & \textbf{\scalebox{.8}{Leipzig Corpora (2016)}} & \textbf{soWaC16 (2016)} & \textbf{Total}\tabularnewline \midrule
gu  &   243K & 531K & 600K & - & 1.36M  \tabularnewline 
kk  &   1M & 7.5M & 1M & - & 9.51M  \tabularnewline
so  &   32K & 123K & - & 1.83M & 1.97M \tabularnewline  \midrule
en  &   843K(gu) 4.44M(kk) 654K(so) & 517K(gu) 5.07M(kk) 1.32M(so) & - & - & 1.36M(gu) 9.51M(kk) 1.97M(so)\tabularnewline \bottomrule 
\end{tabular}
\vspace*{-0.5em}
\caption{Monolingual data sources (with data size in number of sentences). WMT data is obtained from \href{http://data.statmt.org/news-crawl/}{http://data.statmt.org/news-crawl/}. Leipzig corpora is obtained from \href{https://wortschatz.uni-leipzig.de/en/download/}{https://wortschatz.uni-leipzig.de/en/download/}. soWaC16 is obtained from \href{http://habit-project.eu/wiki/SetOfEthiopianWebCorpora}{http://habit-project.eu/wiki/SetOfEthiopianWebCorpora}.}
\vspace{-4mm}
\label{table:table_mono}
\vspace{-0.5em}
\end{table*}



\subsection{Comparable Data Mining} 
\label{sec:comparabledata}


Our comparable data comes from linked Wikipedia pages in different languages obtained using the langlinks from Wikimedia dumps\footnote{\href{https://dumps.wikimedia.org/}{https://dumps.wikimedia.org/}}. To extract sentences from a document, we first translate the source sentence to English using a word dictionary, then we quantify the word overlap using Jaccard similarity score, which is defined as:
\begin{equation*}\label{eq:1}
\small
    Jaccard\_sim(s,t) = \frac{|s \cap t|}{|s \cup t|}
\end{equation*}
We then select pairs that have Jaccard Similarity of at least 0.1. A detailed pseudocode of the extraction process is provided in the Appendix.  We leave out-of-vocabulary words unchanged during the translation process.

To create our word dictionary, we first get a seed dictionary obtained from crowd-sourced and publicly available Panlex's World vocabulary listing\footnote{\href{http://vocab.panlex.org}{http://vocab.panlex.org}}, which we call dict(\textit{Panlex}). Based on this lexicon, we then create a higher coverage dictionary by first training monolingual word embeddings for each language using fastText's skipgram model \cite{bojanowski2017enriching} on the monolingual data of the language (Table \ref{table:table_mono}); then 
learn a linear mapping between the source and target word embeddings with \textit{MUSE} \cite{conneau2017word}, using dict(\textit{Panlex}) as seed translations. Based on this learned mapping, we find translations of up to 200k most frequent words from each language monolingual data by projecting the source language word embedding (gu, kk, and so) to the target language word embedding (en) and taking as translation the target word that has the highest cosine similarity based on Cross-Domain Similarity Local Scaling (CSLS) metric \cite{conneau2017word}, which adjusts cosine similarity values of a word based on the density of the area where its embedding lies. We call this higher coverage dictionary dict(\textit{Projected}) that we use to mine comparable sentences.

\subsection{Comparable Data Evaluation} \label{subsec:mining-eval}
We empirically evaluate the quality of our mined data by comparing our dictionary-based approach 
to a recent state-of-the-art unsupervised bitext retrieval approach 
on the Tatoeba en-kk and en-gu similarity search benchmark \cite{tiedemann-2020-tatoeba}. 

\begin{table}[ht!]
\small
\centering
\renewcommand{\arraystretch}{0.2} 
\begin{tabular}{Z W W W W W W} 
\toprule
\textbf{Method}&\multicolumn{3}{c}{\textbf{en-kk}} & \multicolumn{3}{c}{\textbf{en-gu}}
\\
\cmidrule(lr){2-4}\cmidrule(lr){5-7}
&\multicolumn{1}{c}{\textbf{P}}&\multicolumn{1}{c}{\textbf{R}}&\multicolumn{1}{c}{\textbf{F1}}& \multicolumn{1}{c}{\textbf{P}}&\multicolumn{1}{c}{\textbf{R}}&\multicolumn{1}{c}{\textbf{F1}}\\
\midrule
mBERT + RMSS & 0.25 & 0.24 & 0.24 & 0.21 & 0.2 &0.2 \\ \midrule
dict(\textit{Projected}) +$Jaccard\_sim$ &0.3&0.19&0.23 &0.27&0.14&0.18\\
\bottomrule
\end{tabular}
\caption{Tatoeba bitext mining test results (\textbf{P}recision, \textbf{R}ecall, \textbf{F1}) for kk-en and gu-en pairs, using (1) Ratio Margin-bases Similarity Score (RMSS) calculated on mean-pooled mBert embeddings, (2) Jaccard Similarity of sentence pairs translated with dict(\textit{Projected}). Our simple dictionary-based approach performs closely to mBERT + RMSS in terms of the F1 scores.}
 \vspace{-4mm}
\label{table:comparable_data}
\end{table}

We use our dict(\emph{Projected}) to word-translate the sentences of the kk-en and gu-en Tatoeba datasets, and find closest pairs by computing the Jaccard similarity between the word-translated source and target sentences. 
In parallel, following \citet{keung2020unsupervised}, we convert all source and target sentences of the Tatoeba test sets into embedding vectors with mBERT \cite{devlin-etal-2019-bert}. Given a sentence, we mean-pool embeddings and compute a Ratio Margin-based Similarity Score (RMSS) between each source sentence and its k-nearest target neighbors. Letting $cos(\cdot,\cdot)$ be cosine similarity and NN$_{k}^{src}(x)$ the $k$ nearest neighbors of $x$ in the source embedding space, RMSS$(x,y)$ is:
\[
 \frac{cos(x,y)}{\sum_{z \in \text{NN}^{tgt}_{k}(x)} \frac{cos(x,z)}{2k} + \sum_{z \in \text{NN}^{src}_{k}(y)} \frac{cos(y,z)}{2k}}
\]

Intuitively, RMSS is high when the original source and target pairs are closer compared to their respective neighbors. We set $k=4$. Our simple dictionary-based approach performs closely to mBERT + RMSS as shown in \cref{table:comparable_data} with better precision and lower recall suggesting that, given a pair of sentences, it provides a comparable similarity signal for a pair of sequences. 



\begin{table}[ht]
\small
\resizebox{0.49\textwidth}{!}{
\begin{tabular}{lrrr}
\toprule
                  & en-gu   & en-kk & en-so   \\
                 \midrule
Parallel data (sentences) & 22k & 222k & 52k \\
Mined data (sentences) & 34k & 54k & 4k \\   
Dictionary Entries (words) & 71k & 18k & 2k \\
Syntactic distance & 0.42 & 0.55 & 0.40\\
Character overlap & 0.13 & 0.11 & 0.51\\
Token overlap (BPE) & 0.31 & 0.38 & 0.49\\
Shared WALS typological features & 20 & 3 & 31 \\
\bottomrule
\end{tabular}
}
\caption{Size of our mined comparable data and the parallel training data from the WMT 2019 News Translation Task (for Kazakh and Gujarati) and LORELEI \cite{tracey2019corpus} (for Somali), and the size of the Panlex dictionary that we use. We also give the typological distances between each of the language pairs, quantified using \texttt{lang2vec} vectors \citep{littell2017uriel, malaviya17emnlp}, as well as character, token overlaps (discussed in \S \ref{para: ling_sim}.), and the number of  shared typological features between the languages documented in WALS \cite{wals}. 
}
\vspace{-4mm}
\label{table:supervised_comparison}
\end{table}

\subsection{Data Augmentation by Code-Switching}
\label{data:cs}

Previous works have shown that fine-tuning multilingual models such as multilingual BERT on code-switched data can improve performance for cross-lingual transfer \cite{akyurek2020multi, qin2020cosda}. 
To induce better cross-lingual alignment in our \textit{LM}, we use our dict(\textit{Projected}) to code-switch our \textit{monolingual} source (or target) 
sentences to the target (or source) language. 
We use code-switched sentences that have between 20\% and 50\% of their words translated from their original sentences. We use these collections of code-switched sentences as an ``additional language'' corpus, in addition to the English and foreign language monolingual corpora, to pre-train our LM. 
Despite its imperfect nature, we observe that utilization of such code-switched sentences conditions the LM to the MT-related task of predicting a masked word in a sentence while attending to both the English and foreign language words in the sentence, which results in better performance of the downstream MT task. We note the MLM step that utilizes the code-switched data along with the original data by \emph{MLM$_{cs}$}.
 


\section{Experiments and Results}
The languages we study are Gujarati, Kazakh and Somali. They are spoken by 55M, 22M and 16M speakers worldwide, respectively, and are typologically distant from English (see Table \ref{table:supervised_comparison} for details), and different in terms of writing scripts and alphabets. Additionally, these languages have few parallel but some publicly available comparable and/or monolingual data available, which makes them ideal candidates for low-resource NMT study.


\subsection{Monolingual Data} \label{sec:monodata}
Our monolingual data (\cref{table:table_mono}) are carefully chosen from the same domain of news data and from similar time periods (late 2010s) to mitigate domain discrepancy between source and target languages as per previous research \cite{kim2020and}. 
For English data, we use Wikipedia pages linked to gu, kk and so pages, respectively, 
combined with the randomly down-sampled WMT NewsCrawl corpus so that target and source data are equal in size. 

\subsection{Experimental Setup} \label{subsec:experimentalsetup}
We use WMT 2019 news test set for evaluation of Gujarati $\leftrightarrow$ English and Kazakh $\leftrightarrow$  English. 
We use DARPA's LORELEI \citep{tracey2019corpus} validation and test data sets for Somali.

Because we are interested in simulating a limited compute resource setting, 
instead of training our LMs and MTs for prolonged times, we take into account the average monthly income of each region the language is primarily spoken in (Gujarat for Gujarati, Kazakhstan for Kazakh and Somalia for Somali) and use Amazon AWS EC2 rate as an estimate on how long we should train\footnotemark[1]. 
We assume one month’s worth of average monthly income for \textit{each} of the training steps: \textit{MLM}, \textit{BT+AE}, \textit{BT+MT}, which comes down to 40, 60, and 72 hours on 1 GPU (32GB) each, for gu, so, and kk, respectively. This does not include the time used for lexicon induction or comparable data mining, which are relatively fast (<1 hour each per language). 

With respect to reproducibility, our training configurations, all preprocessing steps and hyperparameters, unless mentioned explicitly, are the ones provided by default in the original XLM repository\footnote{ \href{http://github.com/facebookresearch/XLM}{http://github.com/facebookresearch/XLM}}. Specifically, for every language pair we extract a shared 60k subword vocabulary using Byte-Pair Encoding \cite{sennrich2015neural} with the provided data pre-processing script\footnote{\href{https://github.com/facebookresearch/XLM/blob/main/get-data-nmt.sh}{https://github.com/facebookresearch/XLM/blob/main/get-data-nmt.sh}}. For both the LM and NMT model training we use 1024 as the embedding Layer size, 6 as the number of Transformer layers, 8 as the number of Transformer heads, 0.1 dropout, 0.1 dropout in the attention layer, GELU activation instead of ReLU, and 256 as the sequences' length. We use perplexity (for LM) and en-xx BLEU (for NMT) on validation set as a stopping criterion---stopping when performance does not improve in 10 epochs. For the LM, we use the Adam optimizer with learning rate \textit{lr}=0.0001, 32 as the number of sentences per batch, and 200k number of sentences per epoch. For NMT we use maximum vocabulary size of 200k, Adam optimizer with inverse square root schedule with parameters beta\textsubscript{1}=0.9, beta\textsubscript{2}=0.98, and \textit{lr}=0.0001; and we use a fixed number of 2k words per batch. In addition, for unsupervised NMT, we use the default parameters for the auto-encoding loss, word\_shuffle, word\_dropout, word\_blank, and the auto-encoding coefficient i.e., lambda\_ae\footnote{ \href{https://github.com/facebookresearch/XLM\#train-on-unsupervised-mt-from-a-pre-trained-model}{https://github.com/facebookresearch/XLM\#train-on-unsupervised-mt-from-a-pre-trained-model}}. 

\begin{table*}[ht]
\small
\centering
\resizebox{\textwidth}{!}{
\begin{tabular}{lllrrrrrr}
\toprule
\textbf{Name} & \textbf{Supervision} & \textbf{Multilinguality} & \textbf{en-gu} & \textbf{gu-en} & \textbf{en-kk}  & \textbf{kk-en}  & \textbf{en-so}  & \textbf{so-en}   \\ \midrule
\multicolumn{9}{l}{\emph{Other methods}} \\ \midrule
Google Translate & Supervised & Multilingual & 31.4 & 26.2 & 23.1 & 28.9 & 22.7 & 27.7 \\
mBART25$^1$ & Supervised & Multilingual & 0.1 & 0.3 & 2.5 & 7.4 & - & - \\ 
Previous System (WMT and others) & Supervised & Bilingual & 28.2$^2$ & 24.9$^3$ & 11.1$^3$ & 30.50$^3$ & - & 25.4$^4$ \\ \midrule
\multicolumn{9}{l}{\emph{Ours w/ XLM, Training Objectives $\downarrow$ (Time-Constrained)}} \\ \midrule
\emph{MLM} + (\emph{BT} + \emph{AE}) & Unsupervised & Bilingual & 1.7 & 1.2 & 1.0 & 1.3 & 8.1 & 7.4 \\ \midrule
(\emph{BT} + \emph{MT$_{c}$}) & Weakly-supervised & Bilingual & 3.7 & 0.9 & 0.6 & 1.2 & 1.2 & 1.2  \\ 
\emph{MLM} + (\emph{BT} + \emph{MT$_{c}$}) & Weakly-supervised & Bilingual & 11.7 & 8.3 & 3.4 & 4.2 & 12.8 & 12.6  \\ 
\emph{MLM} + (\emph{BT} + \emph{AE}) + (\emph{BT} + \emph{MT$_{c}$}) & Weakly-supervised & Bilingual  & 14.2 & 10.4 & 5.2 & 6.9 & 13.8 & 13.7 \\ 
\emph{MLM$_{cs}$} + (\emph{BT} + \emph{AE}) + (\emph{BT} + \emph{MT$_{c}$}) & Weakly-supervised & Bilingual & \underline{15.0} & \underline{11.8}& 5.6 & 7.5 & 14.7 & 13.9  \\ \midrule
(\emph{BT} + \emph{MT})  & Supervised & Bilingual & 3.7 & 1.2 & 1.9 & 3.1 & 20.1 & 23.1 \\
\emph{MLM$_{cs}$} + (\emph{BT} + \emph{AE}) +  (\emph{BT} + \emph{MT}) & Supervised & Bilingual & 13.0 & 7.7 & \textbf{\underline{7.9}} & \textbf{\underline{10.7}} & \textbf{\underline{23.6}} & \textbf{\underline{29.3}} \\ \midrule
\multicolumn{9}{l}{\emph{Ours w/ XLM, Training Objectives $\downarrow$ (Until Convergence)}} \\ \midrule
\emph{MLM} + (\emph{BT} + \emph{AE}) + (\emph{BT} + \emph{MT$_{c}$}) & Weakly-supervised & Bilingual & 15.7 & 13.0 & 4.7 & 6.2 & 14.4 & 14.4 \\
\emph{MLM$_{cs}$} + (\emph{BT} + \emph{AE}) + (\emph{BT} + \emph{MT$_{c}$}) & Weakly-supervised & Bilingual & \textbf{15.9} & \textbf{13.2} & 5.5 & 7.2 & 14.6 & 14.0 \\
\bottomrule
\end{tabular}}
\vspace*{-0.5em}
\caption{BLEU scores for previous supervised and unsupervised results from $^1$\citet{liu2020multilingual}, $^2$\citet{bei-etal-2019-gtcom},  $^3$\citet{li-etal-2019-niutrans} and $^4$\citet{liu2018context} and our weakly-supervised models which leverage comparable data. 
Test and validation sets are from WMT19 for Gujarati and Kazakh  and from \citet{tracey2019corpus} for Somali. \emph{MLM}, \emph{AE}, \emph{BT} and \emph{MT} stand for \textit{MLM}, Auto-Encoding loss, Back Translation loss and Machine Translation loss, respectively. \emph{MT} and \emph{MT$_c$} utilize human-labeled data and comparable data, respectively. \emph{MLM$_{cs}$} utilizes both code-switched and original forms of monolingual data. 
Best results overall are \textbf{bolded} while best results under time-constrained setting are \underline{underlined}. All our models here 
use 1 (32GB) GPU. 
Parentheses in training objectives refer to two simultaneous losses while the ones separated with ``+'' (outside of the parentheses) are used successively. 
}
\vspace{-5mm}
\label{table:benchmark}
\end{table*}

\subsection{Results}

\paragraph{Time-Constrained Setting} In Table \ref{table:benchmark}, we explore BLEU scores for several NMT training configurations. First we provide performances for state-of-the-art, highly engineered, supervised translation technologies. We follow with unsupervised MT which is LM pre-training of XLM with the \emph{MLM} objective, followed by back-translation \emph{BT} and auto-encoding \emph{AE} tasks i.e., \emph{MLM} + (\emph{BT} + \emph{AE})---using the monolingual data in   \cref{table:table_mono}. This results in poor BLEU scores similar to what has been reported in previous works \cite{kim2020and,liu2020multilingual}. 
While adhering to the time constraints specified in \S \ref{subsec:experimentalsetup}, we explore using the proposed \emph{MT$_c$} objective that leverages the mined comparable data in different weakly-supervised settings. We observe that (\emph{BT} + \emph{MT$_c$}) itself challenges \emph{MLM} + (\emph{BT} + \emph{AE}) in en-gu, while \emph{MLM} + (\emph{BT} + \emph{MT$_c$}) outperforms (\emph{BT} + \emph{MT$_c$}). When combined together in \textit{MLM} + (\textit{BT}+\textit{AE}) + (\textit{BT}+\textit{MT$_c$}), they outperform individual employments of these objectives suggesting that pre-training on the related tasks of \emph{MLM} and the unsupervised objectives \emph{BT} and \emph{AE} are pivotal to \emph{MT} training. 
We further show that the best gains are achieved when the combined pipeline is preceded with \emph{MLM$_{cs}$} which makes use of code-switched data. Finally, we show that the intermediate steps of \emph{MLM} + (\textit{BT}+\textit{AE}) are imperative to achieve the best performance even when parallel corpora are available (see the Supervised rows in the Time-Constrained section of \cref{table:benchmark}). The performance of our training curriculum with supervised data is higher than our weakly-supervised approach for Kazakh and Somali as the supervised training data for these languages are much larger than our mined comparable data i.e., 2x and 25x our comparable data for Kazakh and Somali, respectively (Table~\ref{table:supervised_comparison}). It is worth noting that the use of our training curriculum with supervised data, despite its simplicity and time-constrained training, results in a new SOTA MT result on the Somali test set (BLEU scores of 29.3 for so-en).

Although we have included the performances of other SOTA supervised models in \cref{table:benchmark} for completeness, different from our models, these are highly engineered models that use more language-specific pre-processing, and much more parallel data. For example, \citet{li-etal-2019-niutrans} leverage related high-resource pivot languages to get more parallel data for Kazakh and Gujarati. Hence, they are not directly comparable to our model, which is a more general approach applied to low resource setting. These approaches are however orthogonal to ours and can be combined with ours to further improve performance e.g., using better, more language-specific tokenization \cite{sanchez2019universitat} might improve our performance for highly inflected languages. 

\paragraph{Training until Convergence} In the bottom section of \cref{table:benchmark}, we showcase how prolonged times of each training step adds an additional boost in translation performance. Comparing \emph{MLM}/\emph{MLM$_{cs}$} + (\textit{BT}+\textit{AE}) + (\textit{BT}+\textit{MT$_c$}) in the last rows to their equivalent counterparts in the middle section where we assumed compute budget is limited reveals that longer training can result in improvements of up to 2.6 BLEU points.

\paragraph{Training with more GPUs} Previous research demonstrates that LM pre-training significantly improves performance in many downstream NLP tasks, including NMT \cite{howard2018universal, lample2018unsupervised} and that LMs benefit from large batch sizes during pre-training \cite{liu2019roberta}. In our experiments, we observe that increasing the availability of compute resources (practically enabling large batch sizes) affect the LM perplexity as well as translation, even when data is scarce. Note that while using gradient accumulation one can mimic a larger number of GPUs, it can quickly become prohibitive time-wise, especially considering the already prolonged times required to train transformer-based LMs. 

In in \cref{table:table2} we provide BLEU scores for different numbers of compute settings for both Kazakh and Somali. 
For 4 GPUs experimental setup (we use NVIDIA V100 32GB GPUs), we increase the batch size to 64 (from 32 with 1 GPU) for LM pre-training, and the tokens per batch to 3k (from 2k with 1 GPU) for MT fine-tuning, per GPU. The results support the hypothesis that enhanced compute resource bears significant potential to boost performance. 
Hence, accounting for the compute resources is indispensable when populating the leaderboards with sophisticated solutions.

In our case, notably, increasing the number of GPUs improves the unsupervised translation scores consistently in general, if not dramatically (Table \ref{table:table2}). For Somali, BLEU scores almost double in both directions, rising from 8.5 to 14.8 for en-so and from 8.0 to 14.8 for so-en. Kazakh proved to be a more challenging case for low-resource NMT (see the following discussion), 
nonetheless, simply utilizing more resources even for Kazakh results in significant improvements in \cref{table:table2} in the unsupervised setting in any translation direction while remaining on par on average for weakly-supervised setting that leverages comparable data. 

\begin{table}[ht]
\small
\resizebox{0.46\textwidth}{!}{
\begin{tabular}{c H H H H}
\toprule
\textbf{GPUs} & \textbf{en-kk} & \textbf{kk-en} & \textbf{en-so} & \textbf{so-en} \\
\hline
\multicolumn{5}{l}{\emph{Unsupervised NMT: MLM + (BT+AE)}} \\ \hline
 1 &  1.1 & 1.6 & 8.5 & 8.0 \\
 4 & \textbf{2.9}& \textbf{3.9} & \textbf{14.8} & \textbf{14.8}\\
\hline \hline
\multicolumn{5}{l}{\emph{Weakly-supervised NMT: MLM + (BT+AE) + (BT+MT$_c$)}} \\ \hline
 1 & \textbf{6.2} & 4.9 & 14.4 & 14.4 \\
 4 & 4.3 & \textbf{6.4}  & \textbf{16.2} & \textbf{15.7}\\
  \bottomrule
\end{tabular}}
\caption{BLEU scores using different number of 32GB V100 GPUs. \emph{MT$_c$} objective uses our comparable data mined using our dictionary (\S \ref{sec:comparabledata}). 
Tokens per batch are set at 3k, which is larger than our basic 1 GPU (32GB) setup (\S \ref{subsec:experimentalsetup}). Best results are bolded.}
 \vspace{-4mm}
\label{table:table2}
\end{table}


\paragraph{Assessing the Effect of Linguistic Similarity}\label{para: ling_sim}
We also measure the similarity between languages in the language pairs we work with, and examine the relationship between linguistic similarity and MT performance. We quantify linguistic similarity in a handful of ways. First, we use \texttt{lang2vec} vectors from \citet{littell2017uriel, malaviya17emnlp}, which give continuous representations of the linguistic properties of a language, as gathered from online databases. We use syntax vectors from a toolkit\footnote{\href{https://github.com/antonisa/lang2vec}{https://github.com/antonisa/lang2vec}}, to compute linguistic distances as cosine distances. Second, we compute the character-level and token-level (BPE subword vocabulary) overlaps between the two languages in each pair, following \citet{jones2021massively}. 
These two metrics quantify the degree of surface-level (i.e. textual) overlap between  two languages. 

These five metrics are given in Table \ref{table:supervised_comparison}. English and Kazakh are shown to be furthest apart syntactically ($d=0.55$), compared to English and Gujarati ($d=0.42$) and English and Somali ($d=0.40$). This could in part explain the relatively poorer performance of en-kk. English and Kazakh also have the lowest character overlap ($s = 0.11$) of the language pairs—even lower than English and Gujarati, which are also written in different scripts. Kazakh also has the lowest number of shared typological features (\textit{3}) with English in comparison to Gujarati (\textit{20}) or Somali (\textit{31}). In comparison, languages like German and French have \textit{103} and \textit{101} shared typological features respectively with English documented in WALS \cite{wals}. 

\section{Conclusion and Future Work}
In this work we explore a wide range of techniques from the NMT toolkit  for low-resource languages and propose a training curriculum that is effective under low-data and low-compute settings. Among those, effective utilization of comparable data with the correct succession of training objectives results in substantial gains. Despite the fact that the three languages we examine here are low-resource, we demonstrate that a simple bitext mining technique yielded quality comparable corpora. While using lexical translations for parallel corpora extraction has a long history in MT \cite{resnik2003web}, it is under-explored for NMT of low-resource languages. 
We also observe that improvement with using comparable data depends on the size of the comparable data---for Gujarati, this results in significant improvement (since the mined comparable data is more than 3 times the size of the supervised data) while for Somali supervised results are still better (since the supervised data is 25x the size of the mined comparable data). Kazakh BLEU percentages are lower compared to those of Somali and Gujarati. We believe this is due to the morphologically more complex structure of Kazakh \citep{maryland}. Yet, the introduction of comparable data substantially increases the weakly-supervised BLEU scores for Kazakh from unsupervised, by 4.6 and 6.2 points, even without additional compute power.

We further compare the similarity signal provided by the mining technique used in this work to a state-of-the-art unsupervised cross-lingual LM's alignment ability in \S\ref{subsec:mining-eval} showing that it's comparable. 
Another useful trick we proved useful for low-resource NMT is the use of code-switched corpora \textit{during} LM pre-training---yielding improvements of MT performance 
(\cref{table:benchmark}). Additionally, given that in none of our experiments do we assume a related high-resource language to aid in translation, or use any parallel sequences, we set the ground for similar analyses and extension of our approaches to other low-resource languages.

Lastly, in conducting our experiments, we were careful to be faithful to the \textit{potentially} resource-constrained settings of the researchers working on low-resource languages. Specifically, controlling for training time and resources shed light on their significant yet unaccounted benefits in improving transformer-based NMT. Such benefits scale up to an impressive 6.8 BLEU points (so-en \cref{table:table2}). While our work focuses more on the optimal training pipeline than budget allocation, a complete analysis around compute budget from a practitioner's perspective 
for different languages with different characteristics, size and quality of training data is intriguing as a future work direction. 
We are also interested in exploring other non-Wikipedia sources for mining comparable sentences, such as international news sites (e.g. Voice of America) and other ways of obtaining lexical translations \cite{conneau2017word,irvine2017comprehensive, artetxe-etal-2017-learning,hewitt2018learning}.

\newpage
\bibliography{anthology,custom}
\bibliographystyle{acl_natbib}

\clearpage
\onecolumn
\appendix
\section{Appendix}

\begin{table*}[ht]
\centering
\begin{adjustbox}{max width=\textwidth}
\begin{tabular}{lllll}
\toprule
\multirow{1}{*}{\rotatebox[origin=c]{90}{{\textsc{dict}}}}
\multirow{1}{*}{\rotatebox[origin=c]{90}{\scalebox{.4}{(\textsc{projected)}}}}
&\textbf{gu} & \includegraphics[scale=0.5]{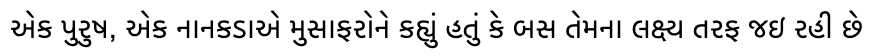}     (A man, a minor, told the passengers that the bus was heading towards their destination.)  \\

&\textbf{en}            &    One of the men, identified as minor, had called for passengers telling them that the bus was going towards their destination.                                        \\ \midrule\midrule

\multirow{1}{*}{\rotatebox[origin=c]{90}{{\textsc{dict}}}}
\multirow{1}{*}{\rotatebox[origin=c]{90}{\scalebox{.4}{\textsc{(projected)}}}}
&\textbf{kk}& \includegraphics[scale=0.5]{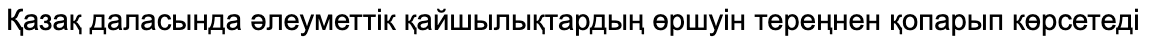}         (It shows the escalation of social conflicts in the Kazakh steppes.)             \\
&\textbf{en}          &   The splash of social contradictions in the Kazakh steppe is shown from its depth After the Tsar’s dethronement the Provisional Government came instead.\\ \midrule \midrule

\multirow{1}{*}{\rotatebox[origin=c]{90}{{\textsc{dict}}}}
\multirow{1}{*}{\rotatebox[origin=c]{90}{\scalebox{.4}{\textsc{(projected)}}}}
&\textbf{so}& Ballankii ugu dambeeyay ee machadka wuxuu ahaa madaxa xafiiska AfDB ee Beeraha iyo Agro-Industry.         (The last appointment of the institute was the head of the AfDB office of Agriculture and Agro-Industry.)             \\
&\textbf{en}          &   His last appointment at the institution was as the head of the AfDB's Department of Agriculture and Agro-Industry.\\
\bottomrule
\end{tabular}
\end{adjustbox}
\vspace*{-0.5em}
\caption{Sample comparable sentence pairs in Gujarati, Kazakh and Somali. Translations in parentheses are obtained using Google Translate which may be imperfect.}
\vspace{-2mm}
\label{table:samples}
\vspace{-0.5em}
\end{table*}

\begin{algorithm}[!htb]
\footnotesize
    \caption{Dictionary based mining}
    \SetAlgoLined
    \SetKwInOut{Input}{Given}
    \Input{$\mathcal{X}, \mathcal{Y}, \mathcal{D}, T$}
    $\mathcal{X}$: Documents in source language\;
    $\mathcal{Y}$: Documents in target language corresponding to $\mathcal{X}$\;
    $\mathcal{D}$: Word Dictionary from source to target language\;
    $T$: Threshold for score\;
    $\mathcal{P} = \{\}$\;
    \For{$x,y \in (\mathcal{X},\mathcal{Y})$}{
        \For{$s \in x$}{
            $best = 0$; $pair = None$\;
            \For{$j=1$ to $len(s)$}{ 
                // Iterate through words\\
                // Translation saved in variable $\mathcal{T}_s$\\
                $\mathcal{T}_{s_j} = \mathcal{D}(s_j)$\;
            }
            \For{$t \in y$}{
                $score = Jaccard\_sim(\mathcal{T}_s,t)$; \\
                // Eq. 1 \\ 
                \If{$score > best \And score > T$} {
                    $best = score$; $pair = (s,t)$\;
                }
            }
            $\mathcal{P}.insert(pair)$\;
        }
    }
    \Return{$\mathcal{P}$}
    \label{algo:mining}
\end{algorithm}

\subsection{Online Code Length Evaluation}
Besides evaluating a model's performance as its ability to generalize to new inputs,  \citet{yogatama2019learning} suggest using the notion of \textbf{(online) codelength} to measure how quickly a model learns a new task, in terms of number of examples seen while training to achieve a certain accuracy.
The codelength $l(A)$ (in bits) of a model with parameters \textbf{W} on a dataset $A = \{(x_i,y_i)\}^N_{i=1}$, which consists of N examples and split into M increasing subsets $S_{1}, S_{2}, ..., S_{M}$, with $S_{M} = A$, 
is defined as: 
\begin{equation}\label{eq:3}
{|S|\textsubscript{1}\log_2|y| - \sum_{i=2}^{N}\log_2 p(y_{S_i}|x_{S_i};\boldsymbol{\hat{W}}_{S_{i-1}})}
\end{equation}
where $|y|$ is the number of possible classes in the data which in our case is the size of vocabulary. This translates into evaluating the model performance i.e., its loss on every subset. As can be seen in \cref{eq:3}, the first term is the initial loss on the first subset where the model just makes a uniform random prediction over the labels. The model that performs well with a limited number of training examples will be rewarded by having a shorter codelength. 


Using the supervised WMT training data for Gujarati split into 3 increasing subsets: 6k, 12k, and 18k sentences, we use our proposed curriculum to train NMT on these increasing training data subsets. We evaluate MT performance on every subset and measure the difference in the online code-lengths between an NMT model trained with our proposed curriculum: \emph{MLM} + (\emph{BT} + \emph{AE}) + (\emph{BT} + \emph{MT}) and a baseline  model trained from scratch: (\emph{BT} + \emph{MT}) on the same architecture and on the same training subsets. For English$\rightarrow$Gujarati, we obtain codelengths of 2517 kbits for the baseline and a much lower 1866 kbits for our model (\cref{table:online_codelength}). Similarly, for Gujarati$\rightarrow$English, we obtain codelengths of 2008 kbits for the baseline and a much lower 1660 kbits for our model\footnote{Note that codelengths are generally higher for NMT than normal classification task since for each example, loss is summed over words in sequences in NMT.}. We also observe that these codelengths correlate with BLEU evaluation: our model that has lower codelengths has higher (final) BLEU scores on the 18k training subset than the baseline (+5.7 for en-gu and +3.8 for gu-en), which is consistent with the findings in \citet{blier2018description}. This suggests that models pre-trained on related tasks such as language modeling and unsupervised NMT can significantly outperform the non-pre-trained ones. One reason can be that pre-training on related NMT task such as unsupervised NMT is analogous to learning what an MT task is. Even though it is learning from different datasets--monolingual instead of bilingual, it can already train the final layer of the decoder so that when used in the first subset, the model can already predict translation non uniformly, thus reducing the initial codelengths (537 instead of 1144 kbits for en-gu and 498 instead of 817 kbits for gu-en in \cref{table:online_codelength}).

\begin{table*}[ht]
\small
\centering
\resizebox{\textwidth}{!}{
\begin{tabular}{llrrrrr}
\toprule
\textbf{Model Name w/ XLM, Training Objectives } & \textbf{Supervision} & \textbf{initial} & \textbf{6k} & \textbf{12k}  & \textbf{18k}  & \textbf{online codelength} \\\midrule
\multicolumn{7}{l}{\emph{en-gu}} \\ \midrule
(\emph{BT} + \emph{MT})  & Supervised & 1143.6 & 484.9 & 472.5 & 416.3 & 2517.3\\
\emph{MLM} + (\emph{BT} + \emph{AE}) + (\emph{BT} + \emph{MT$_{c}$}) & Supervised & 536.9 & 481.8 & 447.5 & 399.8 & 1866.0\\  \midrule
\multicolumn{7}{l}{\emph{gu-en}} \\ \midrule
(\emph{BT} + \emph{MT})  & Supervised & 816.5 & 424.7  & 406.8 & 360.5  & 2008.4\\
\emph{MLM} + (\emph{BT} + \emph{AE}) + (\emph{BT} + \emph{MT$_{c}$}) & Supervised & 497.8 & 419.6  & 394.7 & 348.1  & 1660.2\\ 
\bottomrule
\end{tabular}}
\vspace*{-0.5em}
\caption{Online Codelength Results in kbits for our Gujarati models.}
\vspace{-2mm}
\label{table:online_codelength}
\vspace{-0.5em}
\end{table*}

\end{document}